\documentclass[conference, a4paper, 10pt]{IEEEtran}
\IEEEoverridecommandlockouts
\usepackage{cite}
\usepackage{amsmath,amssymb,amsfonts}
\usepackage{algorithmic}
\usepackage{graphicx}
\usepackage{textcomp}
\usepackage{xcolor}
\def\BibTeX{{\rm B\kern-.05em{\sc i\kern-.025em b}\kern-.08em
    T\kern-.1667em\lower.7ex\hbox{E}\kern-.125emX}}

\usepackage{enumitem}
\usepackage{bbding}
\usepackage{pifont}
\usepackage{algorithm}
\usepackage{algorithmic}
\usepackage{booktabs}
\usepackage{multirow}
\usepackage[super]{nth}
\usepackage{tikz}

\newcommand*\circledB[1]{\tikz[baseline=(char.base)]{
            \node[shape=circle,fill,inner sep=0.2pt] (char) {\textcolor{white}{#1}};}}
\usetikzlibrary{tikzmark}
\tikzset{circledColor/.style={circle,draw,inner sep=0.1em,line width=0.04em}}
\usepackage{fancyhdr}
\fancypagestyle{firstpage}{
  \fancyhf{}
  \chead{Accepted at the International Joint Conference on Neural Networks (IJCNN), June 30th - July 5th, 2025 in Rome, Italy.}
  \cfoot{\thepage}
}

\begin{document}

\title{QSViT: A Methodology for Quantizing \\ Spiking Vision Transformers
\vspace{-0.2cm}
}

\author{
\IEEEauthorblockN{Rachmad Vidya Wicaksana Putra, Saad Iftikhar, Muhammad Shafique}
\IEEEauthorblockA{\textit{eBrain Lab, New York University (NYU) Abu Dhabi, Abu Dhabi, UAE} \\
\{rachmad.putra, si2356, muhammad.shafique\}@nyu.edu}
\vspace{-0.8cm}
}

\maketitle
\pagestyle{plain}
\thispagestyle{firstpage}

%%%%%%%%%%%%%%%%%%%%%%%%%%%%%%%%%%%%%%%%%%%%%%%%%%%%%%%%%%%%%%%%
%%%%%%%%%%%%%%%%%%%%%%%%%%%%%%%%%%%%%%%%%%%%%%%%%%%%%%%%%%%%%%%%
\begin{abstract}
Vision Transformer (ViT)-based models have shown state-of-the-art performance (e.g., accuracy) in vision-based AI tasks. 
However, realizing their capability in resource-constrained embedded AI systems is challenging due to their inherent large memory footprints and complex computations, thereby incurring high power/energy consumption. 
Recently, Spiking Vision Transformer (SViT)-based models have emerged as alternate low-power ViT networks. 
However, their large memory footprints still hinder their applicability for resource-constrained embedded AI systems. 
Therefore, there is a need for a methodology to compress SViT models without degrading the accuracy significantly. 
To address this, we propose QSViT, a novel design methodology to compress the SViT models through a systematic quantization strategy across different network layers.
To do this, our QSViT employs several key steps: (1) investigating the impact of different precision levels in different network layers, (2) identifying the appropriate base quantization settings for guiding bit precision reduction, (3) performing a guided quantization strategy based on the base settings to select the appropriate quantization setting, and (4) developing an efficient quantized network based on the selected quantization setting. 
The experimental results demonstrate that, our QSViT methodology achieves 22.75\% memory saving and 21.33\% power saving, while also maintaining high accuracy within 2.1\% from that of the original non-quantized SViT model on the ImageNet dataset.
These results highlight the potential of QSViT methodology to pave the way toward the efficient SViT deployments on resource-constrained embedded AI systems.  
\end{abstract}


%%%%%%%%%%%%%%%%%%%%%%%%%%%%%%%%%%%%%%%%%%%%%%%%%%%%%%%%%%%%%%%%
%%%%%%%%%%%%%%%%%%%%%%%%%%%%%%%%%%%%%%%%%%%%%%%%%%%%%%%%%%%%%%%%
\section{Introduction}
\label{Sec_Intro}

\begin{figure}[t]
\centering
\includegraphics[width=\linewidth]{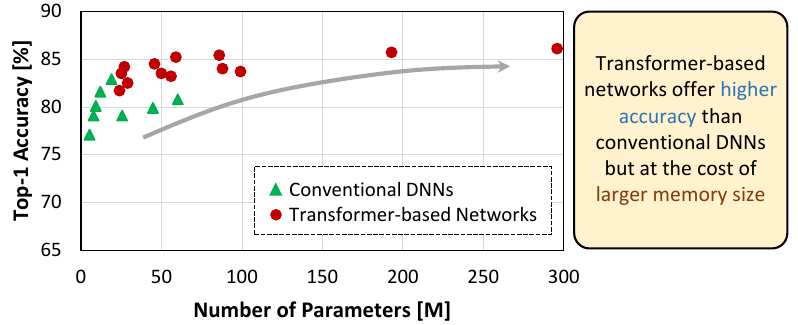}
\vspace{-0.8cm}
\caption{Comparison of the top-1 accuracy for image classification task on the ImageNet dataset~\cite{Ref_Deng_ImageNet_CVPR09} between conventional CNN-based networks and Transformer-based networks, based on the data from~\cite{Ref_Han_SurveyViT_TPAMI22}.} 
\label{Fig_CNNvsTransformers}
\vspace{-0.5cm}
\end{figure}

Recently, Transformer-based networks~\cite{Ref_Vaswani_Attention_NIPS17} have shown state-of-the-art performance in solving various machine learning (ML) problems.
One of the prominent applications is employing Transformers for vision-based tasks (e.g., image classification, object detection, and targeted segmentation), so-called \textit{Vision Transformer (ViT)}~\cite{Ref_Dosovitskiy_Transformers_ICLR21, Ref_Han_SurveyViT_TPAMI22, Ref_Khan_SurveyViT_CSUR22}. 
Currently, ViT models have surpassed the accuracy of state-of-the-art deep neural networks (DNNs) as shown in Fig.~\ref{Fig_CNNvsTransformers}, hence their deployments for embedded AI systems is highly desirable. 
However, realizing their capability in resource-constrained embedded AI systems is very challenging due to their inherent large memory footprints and complex computations, thereby incurring large memory footprint and high power/energy consumption.

In the last decade, spiking neural networks (SNNs) emerged as an energy-efficient alternative for executing neural network (NN) algorithms, due to their bio-inspired sparse event-driven operations~\cite{Ref_Putra_SparkXD_DAC21, Ref_Putra_SpikeDyn_DAC21, Ref_Putra_SoftSNN_DAC22, Ref_Putra_SpiKernel_ESL24}. 
They have been considered in many use-cases that require ultra-low power/energy consumption~\cite{Ref_Bartolozzi_EmbodiedNeuroIntel_Nature22}\cite{Ref_Putra_TopSPark_IROS23}.
Hence, incorporating the effectiveness of Transformers with event-driven SNN computation paradigm potentially improves the energy efficiency of Transformer-based models.
Toward this, recently, Spiking Vision Transformer (SViT) models have been proposed in the literature~\cite{Ref_Zhou_Spikformer_ICLR23, Ref_Yao_SpikeDrivenTransformer_NeurIPS23, Ref_Yao_SpikeDrivenTransformer2_ICLR24}, and becoming the energy-efficient alternatives to the ViT models.
However, their large memory footprints still hinder their applicability for resource-constrained embedded AI systems. 

To reduce the memory requirements of SNN-based models, previous SNN works have proposed different methodologies, including reduction of SNN operations (e.g., neuron elimination~\cite{Ref_Putra_FSpiNN_TCAD20}, stochastic neuron operations~\cite{Ref_Sen_ApproxSNN_DATE17}, and pruning~\cite{Ref_Rathi_PruneQuantizeSNN_TCAD18}), and quantization~\cite{Ref_Rathi_PruneQuantizeSNN_TCAD18, Ref_Sorbaro_OptimSNN_FNINS20, Ref_Zou_MedianQuant_ISCAS20, Ref_Putra_QSpiNN_IJCNN21}.
Among these techniques, quantization is a prominent one since its precision reduction incurs relatively low power/energy overheads. 
However, quantization may also degrade accuracy if it is not performed properly due to information loss~\cite{Ref_Putra_QSpiNN_IJCNN21}.

\textit{\textbf{Targeted Research Problem:} 
If and how can we quantize the Spike-driven Vision Transformer (SViT) model to effectively reduce its memory requirement, while maintaining high accuracy}?
A solution to this problem may enable the efficient SViT deployments on resource-constrained embedded AI systems for diverse application use-cases. 

%%%%%%%%%%%%%%%%%%%%%%%%%%%%%%%%%%%%%%%%%%%%%%%%%%%%%%%%%
\vspace{-0.2cm}
\subsection{State-of-the-art for SViT and Their Limitations}
\label{Sec_Intro_SOTA}
\vspace{-0.1cm}

Currently, several state-of-the-art works on developing SViT models include the Spikformer~\cite{Ref_Zhou_Spikformer_ICLR23} which achieves 74.8\% accuracy with 66.3M parameters, the Spike-Driven Transformer (SDT)~\cite{Ref_Yao_SpikeDrivenTransformer_NeurIPS23} which achieves 77.07\% accuracy with 66.3M parameters, and the Spike-Driven Transformer v2 (SDTv2)~\cite{Ref_Yao_SpikeDrivenTransformer2_ICLR24} which achieves 80\% accuracy with 55.4M parameters; here M refers to million ($10^6$) of counts.
In general, these state-of-the-art mainly focus on the SViT model developments that provide high accuracy, comparable to that of Transformers from DNN domain~\cite{Ref_Zhou_Spikformer_ICLR23, Ref_Yao_SpikeDrivenTransformer_NeurIPS23, Ref_Yao_SpikeDrivenTransformer2_ICLR24}.
However, they still require large memory footprints (about 210MB-250MB), which make it challenging to efficiently deploy them on resource-constrained embedded AI platforms.  
Moreover, \textit{they have not considered employing model compression techniques (such as quantization) to substantially reduce the memory requirements}.

To illustrate the limitations of state-of-the-art and associated research challenges, we perform a case study, which will be discussed further in Section~\ref{Sec_Intro_CaseStudy}. 

%%%%%%%%%%%%%%%%%%%%%%%%%%%%%%%%%%%%%%%%%%%%%%%%%%%%%%%%%
\subsection{Case Study and Associated Research Challenges}
\label{Sec_Intro_CaseStudy}

\begin{figure}[h]
\vspace{-0.2cm}
\centering
\includegraphics[width=\linewidth]{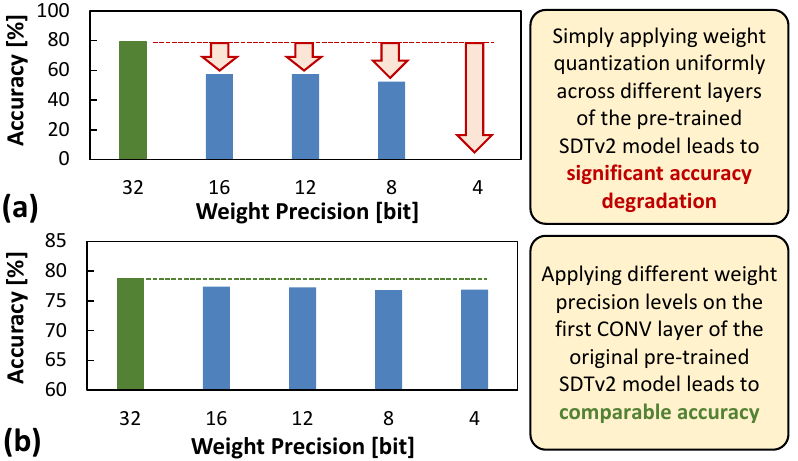}
\vspace{-0.6cm}
\caption{Experimental results after \textbf{(a)} applying different weight precision levels uniformly across different layers of the pre-trained SDTv2 model, and \textbf{(b)} applying different weight precision levels on the first convolutional (CONV) layer of the pre-trained SDTv2 model.
Here, the 32-bit precision denotes the original non-quantized network model.} 
\label{Fig_CaseStudy}
\vspace{-0.4cm}
\end{figure}

This experimental case study aims to investigate the impact of applying post-training quantization (PTQ) on the pre-trained SDTv2 model~\cite{Ref_Yao_SpikeDrivenTransformer2_ICLR24} considering two scenarios. 
\begin{enumerate}[leftmargin=*]
    \item[a)] Applying different weight precision levels (including 16bit, 12bit, 8bit, and 4bit) across network layers. 
    \item[b)] Applying different weight precision levels (including 16bit, 12bit, 8bit, and 4bit) in the same network layer, i.e., the first convolutional (CONV) layer. 
\end{enumerate}
Details of the SDTv2 model are explained in Section~\ref{Sec_Intro_Back_SViT}, while details of the quantization methods and the experimental setup are discussed in Section~\ref{Sec_Intro_Back_Quant} and Section~\ref{Sec_EvalMethod}, respectively.   
The experimental results are presented in Fig.~\ref{Fig_CaseStudy}, from which we draw the following key observations.
\begin{itemize}[leftmargin=*]
    \item Applying quantization across different network layers may significantly degrade the accuracy, since it may remove important information from some layers, as shown in Fig.~\ref{Fig_CaseStudy}(a).
    \item Applying different precision levels in the same network layer may achieve comparable accuracy to the original non-quantized network, indicating that the quantization may not remove important information from the layer; see Fig.~\ref{Fig_CaseStudy}(b). 
    \item Different network layers may have different sensitivity under quantization settings, leading to different accuracy scores. 
\end{itemize}
Although its effective memory saving, quantization may also lead to unacceptable accuracy degradation if it is not carefully performed. 
However, selecting the suitable quantization settings for compressing SViT models is challenging, as different layers may have different sensitivity to the same quantization. 
Therefore, the \textbf{\textit{associated research challenge}} is \textit{how to effectively perform quantization on the SViT models, while offering a good trade-off between memory and accuracy}.

%%%%%%%%%%%%%%%%%%%%%%%%%%%%%%%%%%%%%%%%%%%%%%%%%%%%%%%%%
\subsection{Our Novel Contributions}
\label{Sec_Intro_Novelty}

\begin{figure}[h]
\vspace{-0.2cm}
\centering
\includegraphics[width=\linewidth]{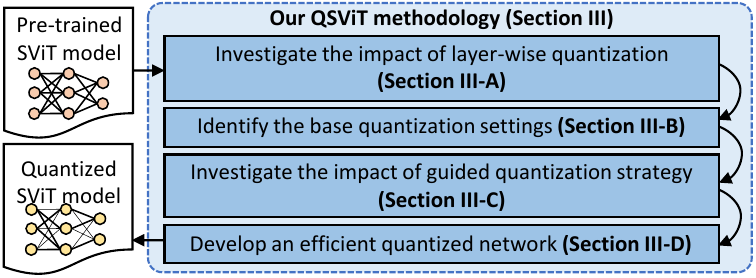}
\vspace{-0.6cm}
\caption{Overview of our novel contributions, highlighted in blue.} 
\label{Fig_Novelty}
\vspace{-0.1cm}
\end{figure}

To address the targeted problem and related challenges, we propose \textit{\textbf{QSViT}, a novel methodology to systematically employ \underline{Q}uantization for \underline{S}piking \underline{Vi}sion \underline{T}ransformers}. 
It employs the following key steps; see an overview Fig.~\ref{Fig_Novelty}.
\begin{itemize}[leftmargin=*]
    \item \textbf{Investigate the impact of layer-wise quantization (Section~\ref{Sec_QSViT_LayerQuant}):} 
    It evaluates the impact of different quantization levels across different network layers to perform a quick assessment for layers' sensitivity. 
    \item \textbf{Identify the base quantization settings (Section~\ref{Sec_QSViT_BaseQuant}):}
    It aims at selecting two quantization settings for providing an effective range for exploring precision level for each network layer, thereby guiding the precision reduction process. 
    \item \textbf{Investigate the impact of guided quantization strategy (Section~\ref{Sec_QSViT_GuidedQuant}):} 
    It aims to explore the impact of reduced precision levels for each layer, leveraging the analysis from the base quantization settings in previous step. 
    \item \textbf{Develop an efficient quantized network (Section~\ref{Sec_QSViT_EfficientQuant}):}
    It leverages the analysis from the guided quantization strategy to select the appropriate quantization setting for developing the compressed network that can maintain high accuracy. 
\end{itemize}

\textbf{Key Results:}
We evaluate our QSViT methodology through PyTorch-based implementation, and then run it on the Nvidia RTX 4090-based multi-GPU machines.
The experimental results show that, our QSViT saves memory footprint by 22.75\% and saves power consumption by 21.33\%, while preserving the accuracy within 2.1\% from the original (non-quantized) network model on the ImageNet dataset~\cite{Ref_Deng_ImageNet_CVPR09}.
These results demonstrate the potential of QSViT methodology to perform effective quantization for SViT models, and thereby paving the way toward the efficient SViT deployments for resource-constrained embedded AI systems.

%%%%%%%%%%%%%%%%%%%%%%%%%%%%%%%%%%%%%%%%%%%%%%%%%%%%%%%%%%%%%%%%
%%%%%%%%%%%%%%%%%%%%%%%%%%%%%%%%%%%%%%%%%%%%%%%%%%%%%%%%%%%%%%%%
\section{Background}
\label{Sec_Back}

%%%%%%%%%%%%%%%%%%%%%%%%%%%%%%%%%%%%%%%%%%%%%%%%%%%%%%%%%
\subsection{Spiking Vision Transformer (SViT)}
\label{Sec_Intro_Back_SViT}

\begin{figure*}[t]
\centering
\includegraphics[width=0.85\linewidth]{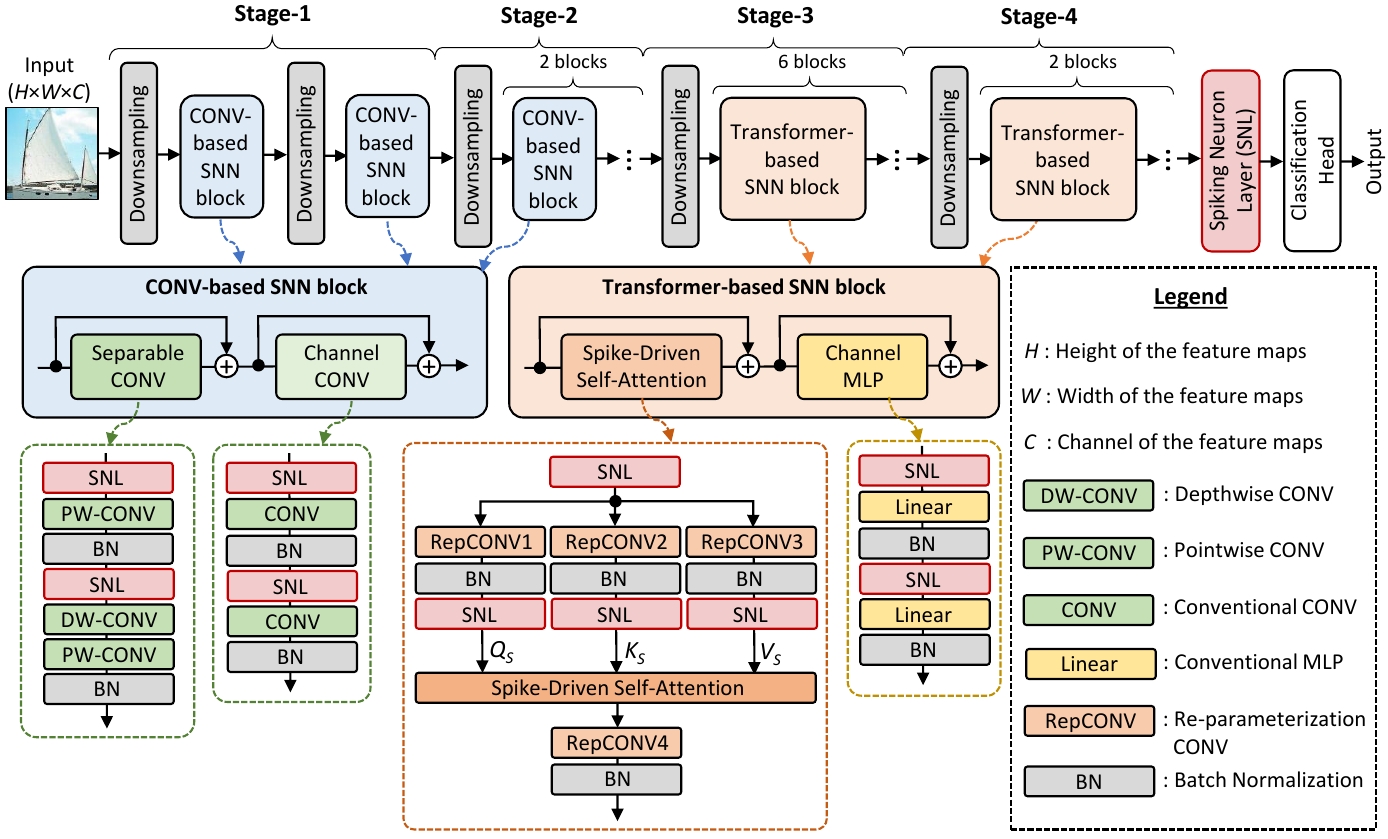}
\vspace{-0.3cm}
\caption{Overview of the SDTv2 network architecture; adapted from studies in~\cite{Ref_Yao_SpikeDrivenTransformer2_ICLR24}. 
There are 4 stages, and each stage has different composition of components.
In Spike-Driven Self-Attention (SDSA), \textit{Q}, \textit{K} and \textit{V} denote the key, query, and value, respectively. 
We refer the detailed configuration of the SDTv2 network architecture to the original paper~\cite{Ref_Yao_SpikeDrivenTransformer2_ICLR24}.} 
\label{Fig_SDTv2}
\vspace{-0.5cm}
\end{figure*}

Currently, state-of-the-art works on developing SViT models include the Spikformer~\cite{Ref_Zhou_Spikformer_ICLR23}, Spike-Driven Transformer (SDT)~\cite{Ref_Yao_SpikeDrivenTransformer_NeurIPS23}, and Spike-Driven Transformer v2 (SDTv2)~\cite{Ref_Yao_SpikeDrivenTransformer2_ICLR24}.
In thsi work, we select the SDTv2 as the reference design, as it offers the state-of-the-art performance compared to other existing works, i.e., by achieving close to 80\% accuracy with 55.4M parameters (i.e., about 210MB considering 32bit weight precision). 
The overview of SDTv2 architecture is presented in Fig.~\ref{Fig_SDTv2}, and discussed in the following.

\smallskip
\textbf{SDTv2 Network architecture:}
At the macro-level, SDTv2 consists of 4 main network stages and a classification head. 
Following are the details each network stage.
\begin{itemize}[leftmargin=*]
    \item \textbf{Stage-1:} 
    It consists of 2 sets of downsampling and CONV-based SNN block, that are composed subsequently.
    The first set considers $(H/2)$$\times$$(W/2)$$\times$$C$ of tokens, while the second set considers $(H/4)$$\times$$(W/4)$$\times$$2C$ of tokens. 
    \item \textbf{Stage-2:}
    It consists of a downsampling block, followed by 2 CONV-based SNN blocks.
    It considers $(H/8)$$\times$$(W/8)$$\times$$4C$ of tokens. 
    \item \textbf{Stage-3:}
    It consists of a downsampling block, followed by 6 Transformer-based SNN blocks.
    This stage considers $(H/16)$$\times$$(W/16)$$\times$$8C$ of tokens. 
    \item \textbf{Stage-4:}
    It consists of a downsampling block, followed by 2 Transformer-based SNN blocks.
    This stage also considers $(H/16)$$\times$$(W/16)$$\times$$10C$ of tokens. 
\end{itemize}
In the CONV-based SNN block, there are Separable CONV and Channel CONV sub-blocks. 
Separable CONV has one Depthwise CONV and two Pointwise CONV modules, while Channel CONV has two conventional CONV modules.
Meanwhile, in the Transformer-based SNN block, there are Spike-Driven Self-Attention (SDSA) and Channel MLP sub-blocks. 
SDSA has four Re-parameterization CONV and an SDSA core modules, while Channel MLP has two conventional Linear MLP modules.

%%%%%%%%%%%%%%%%%%%%%%%%%%%%%%%%%%%%%%%%%%%%%%%%%%%%%%%%%
\subsection{Quantization Method}
\label{Sec_Intro_Back_Quant}

%%%%%%%%%%%%%%%%%%%%%%%
\subsubsection{Scheme}
\label{Sec_Intro_Back_Quant_Schemes}

There are two possible schemes for quantizing neural networks (NN) algorithms, i.e., \textit{Post-Training Quantization (PTQ)}, and \textit{Quantization-aware Training (QAT)}~\cite{Ref_Putra_QSpiNN_IJCNN21}\cite{Ref_Krishnamoorthi_Whitepaper_arXiv18}. 
PTQ quantizes the pre-trained SNN model with the selected precision level, hence resulting in a quantized SNN model for inference.  
In contrast, QAT quantizes an SNN model with the selected precision level during the training phase, thus resulting in a trained SNN model that is already a quantized form and can be used for inference.
In this work, we employ PTQ as it avoids costly training, including the computational time, memory, and power/energy consumption.

\smallskip
%%%%%%%%%%%%%%%%%%%%%%%
\subsubsection{Strategy}
\label{Sec_Intro_Back_Quant_Strategy}

In this work, we consider the integer quantization technique to convert the single-precision floating-point 32bit (FP32) into the desired bit precision ($b$). 
To do this, we first identify the scale of conversion ($S$) by leveraging the range of weight values ($w_{range}=w_{max}-w_{min}$) from the floating-point format and the range of desirable values for quantization ($Q_{range}=Q_{max}-Q_{min}$), as stated in Eq.~\ref{Eq_Scale}.
\begin{equation}
\vspace{-0.2cm}
\small
\begin{split}
   S = \frac{w_{range}}{Q_{range}} \\ 
\end{split}
\label{Eq_Scale}
\end{equation}
$Q_{max}$ and $Q_{min}$ can be obtained by leveraging the desired bit precision ($b$), as stated in Eq.~\ref{Eq_Qvalues}.
\begin{equation}
\small
\begin{split}
   Q_{max}= 2^{(b-1)}-1 \;\;\; \text{and} \;\;\; Q_{min}= -2^{(b-1)}\\
\end{split}
\label{Eq_Qvalues}
\end{equation}
With the scale factor $S$ established, each element of the weight tensor ($\textbf{W}$) can be quantized into an integer representation ($\textbf{W}_{q}$), which is expressed as Eq.~\ref{Eq_Round}.
\begin{equation}
\small
\begin{split}
   \textbf{W}_q[i] = \text{round} \left( \frac{\textbf{W}[i]}{S} \right) \;\;\; \text{with} \;\;\; \textbf{W}_q[i] \in [Q_{min}, Q_{max}]\\
\end{split}
\label{Eq_Round}
\end{equation}
To realize this conversion, we employ the simulated quantization approach to enable fast design exploration, while providing representative results in accuracy and power saving~\cite{Ref_vanBaalen_SimQuantRealPower_CVPR22}. 

%%%%%%%%%%%%%%%%%%%%%%%%%%%%%%%%%%%%%%%%%%%%%%%%%%%%%%%%%%%%%%%%
%%%%%%%%%%%%%%%%%%%%%%%%%%%%%%%%%%%%%%%%%%%%%%%%%%%%%%%%%%%%%%%%
\begin{figure*}[t]
\centering
\includegraphics[width=\linewidth]{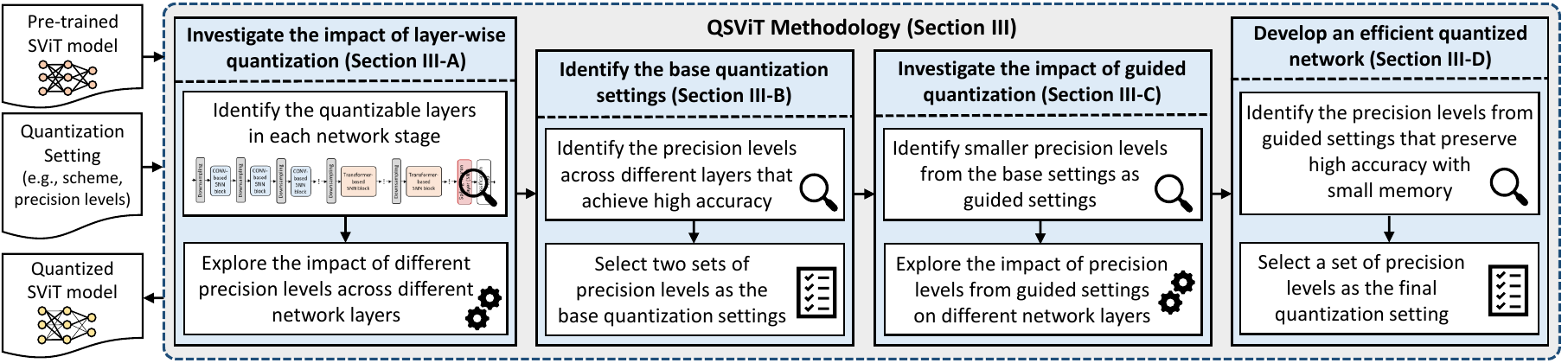}
\vspace{-0.6cm}
\caption{The QSViT methodology, showing its novel design steps in blue boxes.} 
\label{Fig_QSViT}
\vspace{-0.3cm}
\end{figure*}

\section{The QSViT Methodology}
\label{Sec_QSViT}

This methodology aims to address the targeted problem and related research challenges through several key steps, which will be discussed in Section~\ref{Sec_QSViT_LayerQuant} until Section~\ref{Sec_QSViT_EfficientQuant} (see an overview in Fig.~\ref{Fig_QSViT}).

%%%%%%%%%%%%%%%%%%%%%%%%%%%%%%%%%%%%%%%%%%%%%%%%%%%%%%%%%
\subsection{Investigate the Impact of Layer-wise Quantization}
\label{Sec_QSViT_LayerQuant}

This step evaluates the impact of applying different precision levels in each network layer on the accuracy, thereby providing useful insights regarding the sensitivity of each layer under quantization.
It is important, since understanding layers' sensitivity will guide us to devise an effective quantization setting that leads to a quantized model that can preserve high accuracy with a relatively small memory cost. 

To do this, we consider quantizing the state-of-the-art SViT model (i.e., SDTv2~\cite{Ref_Yao_SpikeDrivenTransformer2_ICLR24}), thus providing the state-of-the-art study of quantization for SViT models.
Specifically, we first identify the quantizable layers in each network stage, which are summarized in Table~\ref{Tab_QuantLayer}. 
In each stage, there are several blocks, each encompassing several sub-blocks, and each sub-block consists of several layers. 
For CONV-based and Linear-based layers, quantization can be performed on the weights of CONV, DW-CONV, PW-CONV, RepCONV, and Linear MLP layers. 
Meanwhile, for Transformer-based layers (i.e., SDSA), quantization can be performed on the weights for \textit{Query}, \textit{Key}, and \textit{Value} parameters (i.e., $W_Q$, $W_K$, and $W_V$, respectively). 

\begin{table}[h]
\vspace{-0.2cm}
\caption{List of the quantizable layers in the SDTv2 model.}
\vspace{-0.1cm}
\scriptsize
\begin{tabular}{|c|c|c|c|c|}
\hline
\textbf{Stage} & \textbf{Block} & \textbf{Sub-Block} & \textbf{\begin{tabular}[c]{@{}c@{}}Quantizable \\ Layer\end{tabular}} & \textbf{Quantity} \\ \hline \hline
\multirow{4}{*}{Stage-1} & Downsampling & - & 1 CONV & \multirow{4}{*}{2} \\ \cline{2-4}
 & \multirow{2}{*}{\begin{tabular}[c]{@{}c@{}}CONV-based \\ SNN block\end{tabular}} & Separable CONV & \begin{tabular}[c]{@{}c@{}}1 DW-CONV\\ 2 PW-CONV\end{tabular} &  \\ \cline{3-4}
 &  & Channel CONV & 2 CONV &  \\ \hline
\multirow{4}{*}{Stage-2} & Downsampling & - & 1 CONV & 1 \\ \cline{2-5} 
 & \multirow{2}{*}{\begin{tabular}[c]{@{}c@{}}CONV-based \\ SNN block\end{tabular}} & Separable CONV & \begin{tabular}[c]{@{}c@{}}1 DW-CONV\\ 2 PW-CONV\end{tabular} & \multirow{2}{*}{2} \\ \cline{3-4}
 &  & Channel CONV & 2 CONV &  \\ \hline
\multirow{4}{*}{Stage-3} & Downsampling & - & 1 CONV & 1 \\ \cline{2-5} 
 & \multirow{2}{*}{\begin{tabular}[c]{@{}c@{}}Transformer-based\\ SNN block\end{tabular}} & SDSA Block & \begin{tabular}[c]{@{}c@{}}4 RepCONV\\ 1 SDSA\end{tabular} & \multirow{2}{*}{6} \\ \cline{3-4}
 &  & Channel MLP & 2 Linear &  \\ \hline
\multirow{4}{*}{Stage-4} & Downsampling & - & 1 CONV & 1 \\ \cline{2-5} 
 & \multirow{2}{*}{\begin{tabular}[c]{@{}c@{}}Transformer-based\\ SNN block\end{tabular}} & SDSA Block & \begin{tabular}[c]{@{}c@{}}4 RepCONV\\ 1 SDSA\end{tabular} & \multirow{2}{*}{2} \\ \cline{3-4}
 &  & Channel MLP & 2 Linear &  \\ \hline
\end{tabular}
\label{Tab_QuantLayer}
\end{table} 

After identifying the quantizable layers, we apply different precision levels (i.e., 16-bit, 12-bit, 8-bit, and 4-bit of weight precision) for each layer, and evaluate their impact on the accuracy. 
This proposed exploration strategy is summarized in Alg.~\ref{Alg_LayerQuant}.
Here, quantization process for CONV-based and Linear-based layers are presented in lines 7-10 of Alg.~\ref{Alg_LayerQuant}.
Meanwhile, quantization process for Transformer-based layers are presented in lines 11-16 of Alg.~\ref{Alg_LayerQuant}.
Then, the experimental results for the quantized network models ($qNets$) and the accuracy scores ($acc$) are saved (i.e., lines 17-18 of Alg.~\ref{Alg_LayerQuant}), and will be used for selecting the base quantization settings (discussed in Section~\ref{Sec_QSViT_BaseQuant}).

\begin{algorithm}[h]
\caption{Proposed Layer-wise Quantization Strategy}
\label{Alg_LayerQuant}
\footnotesize
\begin{algorithmic}[1]
\renewcommand{\algorithmicrequire}{\textbf{INPUT:}}
\renewcommand{\algorithmicensure}{\textbf{OUTPUT:}}
\REQUIRE \textbf{(1)} Pre-trained SViT model ($Net$): weights of CONVs or Linear ($Net.W$), weights of Query ($Net.W_Q$), weights of Key ($Net.W_K$), and weights of Value ($Net.W_V$); \\
\textbf{(2)} Number of network layers ($L$); \\
\textbf{(3)} Types of layer: CONV ('conv'), DW-CONV ('dwconv'), PW-CONV ('pwconv'), RepCONV ('repconv'), Linear ('linear'), and SDSA ('sdsa'); \\
\textbf{(4)} Bit precision ($bit$): $bit=[32, 16, 12, 8, 4]$, length of index: $I=5$; \\
\ENSURE \textbf{(1)} Quantized SViT models ($qNets$); \\
\textbf{(2)} Accuracy scores for $qNets$ ($acc$); \\
\smallskip
\textbf{BEGIN} \\
  \textbf{Initialization}: \\
  \STATE $qNet$ = $[]$; \\
  \STATE $qNets$ = $[]$; \\
  \STATE $acc$ = $[]$; \\
  \textbf{Process}: \\
    \FOR{($l=1$; $l<$ ($L$+1); $l$++)}
    \STATE $N0$ = $Net$; \\
    \FOR{($i=1$; $i<$ ($I$+1); $i$++)}
      \IF{($N0[l]$=='conv') or ($N0[l]$=='pwconv') or ($N0[l]$=='pwconv') or ($N0[l]$=='repconv') or ($N0[l]$=='linear')}
        \STATE $qNet$ = quantize($N0[l].W$, $bit[i]$); \\
        \STATE $qNet.bit$ = $bit[i]$; \\
      \ENDIF
      \IF{($N[l]$=='sdsa')}
        \STATE $N1$ = quantize($N0[l].W_Q$, $bit[i]$); \\
        \STATE $N2$ = quantize($N1[l].W_K$, $bit[i]$); \\
        \STATE $qNet$ = quantize($N2[l].W_V$, $bit[i]$); \\
        \STATE $qNet.bit$ = $bit[i]$; \\
      \ENDIF
      \STATE $qNets[l, i]$ = $qNet$; \\ 
      \STATE $acc[l, i]$ = test($qNet$); \\
    \ENDFOR \\
    \ENDFOR \\
  \RETURN $qNets$, $acc$; \\
\textbf{END}
\end{algorithmic} 
\end{algorithm}
\setlength{\textfloatsep}{2pt}

%%%%%%%%%%%%%%%%%%%%%%%%%%%%%%%%%%%%%%%%%%%%%%%%%%%%%%%%%
\vspace{-0.2cm}
\subsection{Identify the Base Quantization Settings}
\label{Sec_QSViT_BaseQuant}

From the previous step in Section~\ref{Sec_QSViT_LayerQuant}, we can identify: (1) which layers that are sensitive to quantization, hence we should avoid quantizing them; and (2) which layers that are non-sensitive to quantization, hence we can perform further study to identify suitable quantization settings (i.e., precision levels) for them. 
We consider layers to be sensitive if their quantization results in more than 5\% accuracy degradation. 

Therefore, in this step, we further identify \textbf{\textit{the base quantization settings}} whose individual precision level each layer leads to acceptable accuracy (i.e., within 5\% accuracy from the original network model), based on the observation from Section~\ref{Sec_QSViT_LayerQuant}. 
Specifically, we try to determine the following.
\begin{enumerate}[leftmargin=*]
    \item The quantization setting with the highest precision levels across network layers ($bSettingH$).  
    \item The quantization setting with the lowest precision levels across network layers ($bSettingL$).
\end{enumerate}
To do this, we devise a selection strategy which is summarized in Alg.~\ref{Alg_BaseSetting}.
We leverage exploration results from Alg.~\ref{Alg_LayerQuant}, and select the quantization settings that achieve accuracy within 5\% from the original non-quantized model (line 6 of Alg.~\ref{Alg_BaseSetting}).
Afterward, the quantization setting with the highest precision levels across network layers will be recorded in $bSettingH$ (lines 7-9 of Alg.~\ref{Alg_BaseSetting}), while the quantization setting with the lowest precision levels across network layers will be recorded in $bSettingL$ (lines 10-16 of Alg.~\ref{Alg_BaseSetting}).

These two settings ($bSettingH$ and $bSettingL$) are important since they can be used for defining the range of precision level exploration, which is required to find the most suitable quantization setting for preserving high accuracy with small memory cost.
Therefore, these two settings will be used for devising the guided quantization process (discussed further in Section~\ref{Sec_QSViT_GuidedQuant}).

\begin{algorithm}[h]
\caption{Selection of the Base Quantization Settings}
\label{Alg_BaseSetting}
% \small
\footnotesize
\begin{algorithmic}[1]
\renewcommand{\algorithmicrequire}{\textbf{INPUT:}}
\renewcommand{\algorithmicensure}{\textbf{OUTPUT:}}
\REQUIRE \textbf{(1)} Pre-trained SViT model ($Net$); \\
\textbf{(2)} Quantized SViT models from $qNets$ and their accuracy scores ($acc$); \\
\textbf{(3)} Number of network layers ($L$); \\
\textbf{(4)} Bit precision ($bit$): $bit=[32, 16, 12, 8, 4]$, length of index: $I=5$; \\
\ENSURE \textbf{(1)} Quantization setting for the highest precision levels across network layers ($bSettingH$); \\
\textbf{(2)} Quantization setting for the lowest precision levels across network layers ($bSettingL$); \\
\smallskip
\textbf{BEGIN} \\
  \textbf{Initialization}: \\
  \STATE $bSettingH[:]$ = 0; \\
  \STATE $bSettingL[:]$ = 0; \\
  \STATE $bAcc$ = test($Net$); \\
  \textbf{Process}: \\
    \FOR{($l=1$; $l<$ ($L$+1); $l$++)}
    \FOR{($i=1$; $i<$ ($I$+1); $i$++)}
      \IF{$acc[l,i]$ $\geq$ $(bAcc$-$5)$}
        \IF{$bit[i]$ $\geq$ $bSettingH[l]$}
          \STATE $bSettingH[l]$ = $bit[i]$;
        \ENDIF
        \IF{$(bitL[l]$==0)}
          \STATE $bSettingL[l]$ = $bit[i]$;
        \ELSE
          \IF{$bit[i]$ $\leq$ $bSettingL[l]$}
            \STATE $bSettingL[l]$ = $bit[i]$;
          \ENDIF
        \ENDIF
      \ENDIF \\
    \ENDFOR \\
    \ENDFOR \\
  \RETURN $bSettingH$, $bSettingL$; \\
\textbf{END}
\end{algorithmic} 
\end{algorithm}
\setlength{\textfloatsep}{4pt}

%%%%%%%%%%%%%%%%%%%%%%%%%%%%%%%%%%%%%%%%%%%%%%%%%%%%%%%%%
\subsection{Investigate the Impact of Guided Quantization}
\label{Sec_QSViT_GuidedQuant}

After determining the base quantization settings in previous step (Section~\ref{Sec_QSViT_BaseQuant}), we have the defined range of precision levels across network layers to be explored further.
To perform the exploration, we propose \textbf{\textit{the guided quantization}} technique, whose mechanism is presented in Alg.~\ref{Alg_GuidedQuant}.

The key idea of our guided quantization is to apply different precision levels within the given range in each network layer (line 7 of Alg.~\ref{Alg_GuidedQuant}), and evaluate their impact on the accuracy (line 9 of Alg.~\ref{Alg_GuidedQuant}). 
Afterward, for each layer, we evaluate if the investigated precision level leads the respective quantized network to achieves less than 5\% accuracy degradation from the original non-quantized network, while providing the lowest precision level (lines 10-11 of Alg.~\ref{Alg_GuidedQuant}). 
If so, then we keep this precision level in $bSettingG$ (line 13 of Alg.~\ref{Alg_GuidedQuant}), meaning that it is considered as the most suitable precision level in the given layer. 
Therefore, the complete $bSettingG$ (line 18 of Alg.~\ref{Alg_GuidedQuant}) stores the selected precision levels for different network layers, and it will be used for developing a quantized network model (discussed further in Section~\ref{Sec_QSViT_EfficientQuant}). 

\begin{algorithm}[h]
\caption{Exploration using the Guided Quantization}
\label{Alg_GuidedQuant}
\footnotesize
\begin{algorithmic}[1]
\renewcommand{\algorithmicrequire}{\textbf{INPUT:}}
\renewcommand{\algorithmicensure}{\textbf{OUTPUT:}}
\REQUIRE \textbf{(1)} Pre-trained SViT model ($Net$) \\
\textbf{(2)} Setting for the highest precision levels across layers ($bSettingH$); Setting for the lowest precision levels across layers ($bSettingL$); \\
\textbf{(3)} Number of network layers ($L$); \\
\textbf{(4)} Bit precision ($bit$): $bit=[32, 16, 12, 8, 4]$, length of index: $I=5$; \\
\ENSURE \textbf{(1)} Quantization setting for the suitable precision levels across network layers ($bSettingG$); \\
\smallskip
\textbf{BEGIN} \\
  \textbf{Initialization}: \\
  \STATE $accStoredG$ = 0; \\
  \STATE $bSettingG$ = $bSettingH$; \\
  \STATE $bAcc$ = test($Net$); \\
  \textbf{Process}: \\
    \FOR{($l=1$; $l<$ ($L$+1); $l$++)}
    \STATE $accStoredG$ = 0; \\
    \STATE $Net_{temp}$ = $Net$; \\
    \FOR{($b=bSettingH[l]$; $b>$ ($bSettingL[l]$-1); $b=b$-4)}
      \STATE $qNet_{temp}$ = $\text{quantize}(Net_{temp}[l], b)$; \
      \STATE $acc_{temp}$ = test($qNet_{temp}$); \\
      \IF{$acc_{temp}$ $\geq$ $(bAcc$-$5)$}
        \IF{$b$ $\leq$ $bSettingG[l]$}
          \STATE $accStoredG$ = $acc_{temp}$; \\
          \STATE $bSettingG[l]$ = $b$; \\
        \ENDIF \\
      \ENDIF \\
    \ENDFOR \\
    \ENDFOR \\
  \RETURN $bSettingG$; \\
\textbf{END}
\end{algorithmic} 
\end{algorithm}
\setlength{\textfloatsep}{4pt}

%%%%%%%%%%%%%%%%%%%%%%%%%%%%%%%%%%%%%%%%%%%%%%%%%%%%%%%%%
\vspace{-0.2cm}
\subsection{Develop an Efficient Quantized Network Model}
\label{Sec_QSViT_EfficientQuant}

After performing the design exploration using the guided quantization process in previous step (Section~\ref{Sec_QSViT_GuidedQuant}), we have the recorded setting of precision levels across network layers (i.e., $bSettinG$). 
This selected quantization setting is then utilized to develop a quantized network model by following Alg.~\ref{Alg_EfficientQuant}.
To do this, we first apply the precision levels from $bSettinG$ layer-by-layer to the original network model (lines 2-4 of Alg.~\ref{Alg_EfficientQuant}).
Afterward, we evaluate the quantized network model (line 5 of Alg.~\ref{Alg_EfficientQuant}). 
Here, it is expected that the developed quantized network can maintain accuracy within 5\% accuracy degradation from the original non-quantized network model with significant memory footprint reduction.

\begin{algorithm}[h]
\caption{Development of the Efficient Quantized Network}
\label{Alg_EfficientQuant}
\footnotesize
\begin{algorithmic}[1]
\renewcommand{\algorithmicrequire}{\textbf{INPUT:}}
\renewcommand{\algorithmicensure}{\textbf{OUTPUT:}}
\REQUIRE \textbf{(1)} Pre-trained SViT model ($Net$) \\
\textbf{(2)} Quantization setting for the suitable precision levels across network layers ($bSettingG$); \\
\textbf{(3)} Number of network layers ($L$); \\
\ENSURE \textbf{(1)} Final quantized network model ($qNet$); \\
\textbf{(2)} Final accuracy ($qAcc$); \\
\smallskip
\textbf{BEGIN} \\
  \textbf{Initialization}: \\
  \STATE $qNet$ = $[]$; \\
  \textbf{Process}: \\
    \FOR{($l=1$; $l<$ ($L$+1); $l$++)}
      \STATE $qNet$ = $\text{quantize}(Net[l], bSettingG[l])$; \
    \ENDFOR \\
  \STATE $qAcc$ = test($qNet$); \\
  \RETURN $qNet$, $qAcc$; \\
\textbf{END}
\end{algorithmic} 
\end{algorithm}
\setlength{\textfloatsep}{4pt}

%%%%%%%%%%%%%%%%%%%%%%%%%%%%%%%%%%%%%%%%%%%%%%%%%%%%%%%%%%%%%%%%
%%%%%%%%%%%%%%%%%%%%%%%%%%%%%%%%%%%%%%%%%%%%%%%%%%%%%%%%%%%%%%%%
\section{Evaluation Methodology}
\label{Sec_EvalMethod}

\begin{figure}[h]
\vspace{-0.2cm}
\centering
\includegraphics[width=\linewidth]{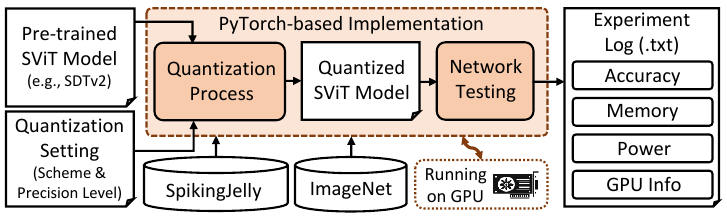}
\vspace{-0.6cm}
\caption{Experimental setup and tools for evaluating our QSViT methodology.} 
\label{Fig_ExpSetup}
\vspace{-0.2cm}
\end{figure}

Fig.~\ref{Fig_ExpSetup} illustrates the experimental setup used to evaluate our QSViT methodology. 
Specifically, we employ a PyTorch-based implementation leveraging the SpikingJelly library~\cite{Ref_Fang_SpikingJelly_SciAdv23}, and run it on the Nvidia RTX 4090-based multi-GPU machines.
Such an implementation takes a pre-trained SViT model and a quantization setting (e.g., scheme and precision level) as inputs, while considering the ImageNet dataset~\cite{Ref_Deng_ImageNet_CVPR09}.
Outputs of the experiments mainly include accuracy, memory, and power consumption of running the quantized SViT model.

\textbf{Comparison Partner:}
We use the state-of-the-art SDTv2 model from the work of~\cite{Ref_Yao_SpikeDrivenTransformer2_ICLR24} as the reference network for the pre-trained SViT model, whose architecture is illustrated in Fig.~\ref{Fig_SDTv2}. 
Therefore, the main comparison partner is also the original (non-quantized) SDTv2 model. 
Note, we run the SDTv2 model using its open-source codes in the same GPU machines to provide fair comparison, and we obtain 78.9\% accuracy, which is slightly different from the data reported in the original paper (i.e., 80\% accuracy). 

\textbf{Quantization:}
For the design space exploration, we consider several quantization levels, including the 16-bit, 12-bit, 8-bit, and 4-bit precision. 
For all experiments, we use the PTQ scheme to quantize the pre-trained SViT model.

%%%%%%%%%%%%%%%%%%%%%%%%%%%%%%%%%%%%%%%%%%%%%%%%%%%%%%%%%%%%%%%%
%%%%%%%%%%%%%%%%%%%%%%%%%%%%%%%%%%%%%%%%%%%%%%%%%%%%%%%%%%%%%%%%
\section{Results and Discussion}
\label{Sec_Results}

%%%%%%%%%%%%%%%%%%%%%%%%%%%%%%%%%%%%%%%%%%%%%%%%%%%%%%%%%
\subsection{Layers' Sensitivity Analysis under Quantization}
\label{Sec_Results_LayerQuant}

\begin{figure*}[t]
\vspace{-0.3cm}
\centering
\includegraphics[width=0.95\linewidth]{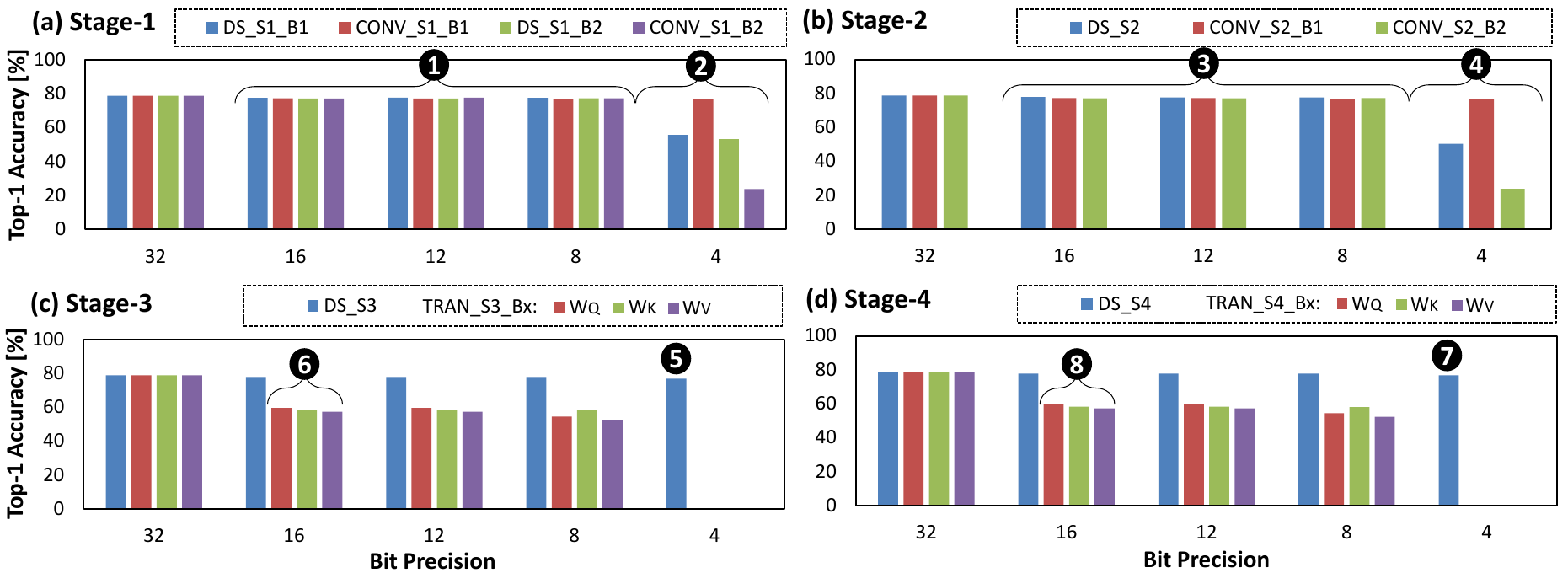}
\vspace{-0.2cm}
\caption{Experimental results for investigating layers' sensitivity considering quantization in \textbf{(a)} Stage-1, \textbf{(b)} Stage-2, \textbf{(c)} Stage-3, and \textbf{(d)} Stage-4. Here, ``DS\_Sm\_Bn'' denotes the Downsampling Block-n in Stage-m, ``CONV\_Sm\_Bn'' denotes the CONV Block-n in Stage-m, while ``TRAN\_Sm\_Bx'' denotes all Transformer blocks in Stage-m.} 
\label{Fig_Sensitivity}
\vspace{-0.4cm}
\end{figure*}

In the first step of our QSViT methodology, we investigate the impact of layer-wise quantization on the accuracy, whose experimental results are shown in Fig.~\ref{Fig_Sensitivity}(a), Fig.~\ref{Fig_Sensitivity}(b), Fig.~\ref{Fig_Sensitivity}(c), and Fig.~\ref{Fig_Sensitivity}(d) for quantization in Stage-1, Stage-2, Stage-3, and Stage-4, respectively.
These results are presented in block-wise since it often provides more intuitive and meaningful information. 
For instance, separate DS-CONV and PW-CONV layers typically serve to reduce the computational cost of a single large CONV layer while providing the same feature extraction functionality, thus understanding the impact of quantization on a single large CONV layer is usually more intuitive.  
Following are our key observations from the results.

\textbf{Stage-1:} 
Quantization with at least 8-bit precision level still can maintain high accuracy, comparable to that of the original non-quantized model (32-bit precision); see~\circledB{1}.
The reason is that, these precision levels maintain important information for meaningful feature extraction.
Meanwhile, 4-bit precision significantly decreases accuracy when applied on the Downsampling blocks (i.e., DS\_S1\_B1 and DS\_S1\_B2) and on the last CONV block (i.e., CONV\_S1\_B2); see~\circledB{2}. 
It indicates that, 4-bit precision level suffers from losing important information, hence hindering from meaningful feature extraction.
Interestingly, 4-bit precision for CONV\_S1\_B1 also does not significantly decrease accuracy, showing its low sensitivity to quantization.
These results highlight the potential of applying different precision levels on different blocks in Stage-1.

\textbf{Stage-2:} 
Quantization with at least 8-bit precision level also maintains high accuracy, comparable to that of the original non-quantized model, as indicated by~\circledB{3}.
The reason is that, these precision levels still maintain important information for meaningful feature extraction.
Meanwhile, 4-bit precision significantly decreases accuracy when applied on the Downsampling blocks (i.e., DS\_S2) and on the last CONV block (i.e., CONV\_S2\_B2), as indicated by~\circledB{4}. 
This precision level losses much of important information, hence hindering from meaningful feature extraction.
Interestingly, 4-bit precision for CONV\_S2\_B1 also does not significantly decrease accuracy, showing its low sensitivity to quantization.
These results also highlight the potential of applying different precision levels on different blocks in Stage-2.

\textbf{Stage-3:} 
For DS\_S3, quantization can be safely performed using 4-bit precision, as this Downsampling block does not suffer from information loss during bitwidth reduction; see~\circledB{5}.
Meanwhile, for all Transformer-based SNN blocks, quantizing the weights for Query ($W_Q$), Key ($W_K$), and Value ($W_V$) to 16-bit precision already causes a significant accuracy degradation; see~\circledB{6}.
This indicates that the Attention module is very sensitive to quantization, hence keeping the 32-bit precision is the reasonable option.
These results also show the potential of applying different precision levels on different blocks in Stage-3, i.e., for Downsampling and Attention parts.

\textbf{Stage-4:} 
For DS\_S4, quantization can also be safely performed using 4-bit precision, as this block does not suffer from information loss during bitwidth reduction; see~\circledB{7}.
Meanwhile, for all Transformer-based SNN blocks, quantizing the weights for Query $W_Q$, Key $W_K$, and Value $W_V$ to 16-bit precision also causes a notable accuracy degradation; see~\circledB{8}.
This also shows that, the Attention module is very sensitive to quantization, hence keeping the 32-bit precision is the reasonable option.
These results also highlight the potential of applying different precision levels on different blocks in Stage-4, i.e., for Downsampling and Attention parts.

%%%%%%%%%%%%%%%%%%%%%%%%%%%%%%%%%%%%%%%%%%%%%%%%%%%%%%%%%
\subsection{Exploration using the Guided Quantization}
\label{Sec_Results_GuidedQuant}

In the second step of our QSViT methodology, we investigate the base quantization settings that are used for guiding further exploration in finding the most suitable quantization setting. 
By employing Alg.~\ref{Alg_BaseSetting}, we obtain the following settings presented in a format of \textbf{Stage-1} (DS\_S1\_B1, CONV\_S1\_B1, DS\_S1\_B2, CONV\_S1\_B2), \textbf{Stage-2} (DS\_S2, CONV\_S2\_B1, CONV\_S2\_B2), \textbf{Stage-3} (DS\_S3, TRAN\_S3\_B1-B6), \textbf{Stage-4} (DS\_S4, TRAN\_S4\_B1-B2).
\begin{itemize}[leftmargin=*]
    \item $bSettingH$: \textbf{Stage-1} (12, 16, 8, 12), \textbf{Stage-2} (16, 12, 8), \textbf{Stage-3} (8, 32), \textbf{Stage-4} (8, 32).
    \item $bSettingL$: \textbf{Stage-1} (8, 4, 8, 8), \textbf{Stage-2} (8, 4, 8), \textbf{Stage-3} (4, 32), \textbf{Stage-4} (4, 32).
\end{itemize}

\begin{figure}[t]
\centering
\includegraphics[width=\linewidth]{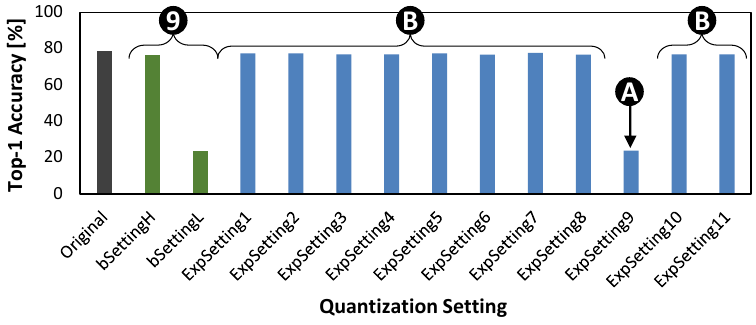}
\vspace{-0.7cm}
\caption{Experimental results of the design exploration using the proposed guided quantization algorithm in Alg.~\ref{Alg_GuidedQuant}.} 
\label{Fig_GuidedExp}
\end{figure}

These settings provide a range that can be used to explore the impact of different precision levels in each layer using the guided quantization algorithm in Alg.~\ref{Alg_GuidedQuant}.
The experimental results are presented in Fig.~\ref{Fig_GuidedExp}, from which we make the following key observations.
\begin{itemize}[leftmargin=*]
    \item $bSettingH$ achieves 76.8\% accuracy, comparable to that of the original model with (78.9\% accuracy); while $bSettingL$ suffers from significant accuracy degradation since it only achieves 23.9\% accuracy; see \circledB{9}.
    \item We also observe that the experimental quantization setting-9 (ExpSetting9) also suffers from significant accuracy degradation like $bSettingL$; see \circledB{A}. We identify that the common aspect that causes ExpSetting9 and $bSettingL$ have low accuracy is due to the employment of 4-bit precision for CONV\_S2\_B1, while having other blocks in reduced precision. Therefore, 4-bit precision should not be used for CONV\_S2\_B1.
    \item The other experimental quantization settings preserve high accuracy scores comparable to that of the original model, as shown by \circledB{B}. 
    From these settings, we select the lowest precision in each layer that leads to high accuracy (i.e., from ExpSetting1-ExpSetting8 and ExpSetting10-ExpSetting11). 
    In result, we select the quantization setting of \textbf{Stage-1} (8, 4, 8, 8), \textbf{Stage-2} (8, 8, 8), \textbf{Stage-3} (4, 32), \textbf{Stage-4} (4, 32).
\end{itemize}

%%%%%%%%%%%%%%%%%%%%%%%%%%%%%%%%%%%%%%%%%%%%%%%%%%%%%%%%%
\subsection{Maintaining High Accuracy}
\label{Sec_Results_HighAccuracy}

From the guided quantization exploration, we obtain the quantization setting of: \textbf{Stage-1} (8, 4, 8, 8), \textbf{Stage-2} (8, 4, 8), \textbf{Stage-3} (4, 32), \textbf{Stage-4} (4, 32). 
Therefore, we then apply this setting to the pre-trained SViT model (i.e., SDTv2) for developing a quantized SViT model, and evaluate it considering the ImageNet dataset. 
The experimental results show that, this quantized SViT model can achieve 76.8\% accuracy, which is within 2.1\% from the original non-quantized one (i.e., 78.9\% accuracy); see Table~\ref{Tab_Comparison}.
These highlight the effectiveness of our QSViT methodology to systematically apply quantization on pre-trained SViT models through carefully-selected precision levels across different network layers, thereby enabling the quantized model to preserve high accuracy.

\begin{table*}[h]
\caption{Comparison of different methods (i.e., ANN, ANN-to-SNN conversion, and SViT) for image classification task considering the ImageNet dataset; based on our experiments and data from studies in~\cite{Ref_Zhou_Spikformer_ICLR23, Ref_Yao_SpikeDrivenTransformer_NeurIPS23, Ref_Yao_SpikeDrivenTransformer2_ICLR24}. Note, results for SDTv2 and QSViT are obtained from our experiments using the same Nvidia RTX 4090 multi-GPU machines.}
\label{Tab_Comparison}
\footnotesize
\centering
\begin{tabular}{|c|c|c|c|c|c|c|c|}
\hline
\textbf{Method} & \textbf{Architecture} & \textbf{\begin{tabular}[c]{@{}c@{}}Spike-Driven\end{tabular}} & \textbf{\begin{tabular}[c]{@{}c@{}}\# Parameters \\ {[}M{]}\end{tabular}} & \textbf{\begin{tabular}[c]{@{}c@{}}Precision \\ {[}bit{]}\end{tabular}} & \textbf{\begin{tabular}[c]{@{}c@{}}Memory Footprint\\ {[}MB{]}\end{tabular}} & \textbf{Timestep} & \textbf{\begin{tabular}[c]{@{}c@{}}Accuracy\\ {[}\%{]}\end{tabular}} \\ \hline \hline
ANN & ViT~\cite{Ref_Dosovitskiy_Transformers_ICLR21} & \ding{55} & 86 & 32 & 328 & 1 & 79.7 \\ \hline
\begin{tabular}[c]{@{}c@{}}ANN-to-SNN\\ Conversion\end{tabular}& \begin{tabular}[c]{@{}c@{}}ResNet-34~\cite{Ref_Rathi_DeepSNN_ICLR20}\\ VGG-16~\cite{Ref_Hu_FastSNN_TPAMI23} \end{tabular} & \begin{tabular}[c]{@{}c@{}} \ding{51} \\ \ding{51} \end{tabular} & \begin{tabular}[c]{@{}c@{}}21.8\\ 138.4\end{tabular} & \begin{tabular}[c]{@{}c@{}}32\\ 32\end{tabular} & \begin{tabular}[c]{@{}c@{}}83\\ 528\end{tabular} & \begin{tabular}[c]{@{}c@{}}250\\ 7\end{tabular} & \begin{tabular}[c]{@{}c@{}}61.5\\ 73.0\end{tabular} \\ \hline
\begin{tabular}[c]{@{}c@{}}Spike-driven ViT\\ (SViT)\end{tabular} & \begin{tabular}[c]{@{}c@{}}Spikformer~\cite{Ref_Zhou_Spikformer_ICLR23} \\ SDT~\cite{Ref_Yao_SpikeDrivenTransformer_NeurIPS23} \\ SDTv2~\cite{Ref_Yao_SpikeDrivenTransformer2_ICLR24}\\ \textbf{QSViT}\end{tabular} & \begin{tabular}[c]{@{}c@{}} \ding{51} \\ \ding{51} \\ \ding{51} \\ \ding{51} \end{tabular} & \begin{tabular}[c]{@{}c@{}}66.34\\ 66.34\\ 55.4\\ \textbf{55.4}\end{tabular} & \begin{tabular}[c]{@{}c@{}}32\\ 32\\ 32\\ \textbf{mixed}\end{tabular} & \begin{tabular}[c]{@{}c@{}}253\\ 253\\ 211\\ \textbf{163}\end{tabular} & \begin{tabular}[c]{@{}c@{}}4\\ 4\\ 4\\ \textbf{4}\end{tabular} & \begin{tabular}[c]{@{}c@{}}74.8\\ 76.3\\ 78.9\\ \textbf{76.8}\end{tabular} \\ \hline
\end{tabular}
\vspace{-0.4cm}
\end{table*}

%%%%%%%%%%%%%%%%%%%%%%%%%%%%%%%%%%%%%%%%%%%%%%%%%%%%%%%%%
\subsection{Memory Footprint Reduction}
\label{Sec_Results_Memory}

After applying the selected quantization setting, we obtain the quantized SViT model which incurs 22.75\% smaller memory footprint than the original non-quantized one.
Specifically, the quantized model requires about 163MB of memory, while the original model requires about 211MB of memory. 
These results highlights the effectiveness of our QSViT methodology to optimize the memory footprint of the SViT models through quantization, while preserving high accuracy.

%%%%%%%%%%%%%%%%%%%%%%%%%%%%%%%%%%%%%%%%%%%%%%%%%%%%%%%%%
\subsection{Power Consumption Saving}
\label{Sec_Results_Power}

After applying the selected quantization setting, we also obtain the quantized SViT model which incurs 21.33\% smaller power consumption than the original non-quantized model.
Specifically, the quantized model consumes about 251.48W of power, while the original model consumes about 319.66W of power memory, when run on the Nvidia RTX 4090-based multi-GPU machines. 
This highlights the effectiveness of our QSViT methodology to reduce data precision in SViT model size, thus leading to smaller computational complexity, and hence resulting in reduced dynamic power consumption.

%%%%%%%%%%%%%%%%%%%%%%%%%%%%%%%%%%%%%%%%%%%%%%%%%%%%%%%%%%%%%%%%
%%%%%%%%%%%%%%%%%%%%%%%%%%%%%%%%%%%%%%%%%%%%%%%%%%%%%%%%%%%%%%%%
\section{Conclusion}
\label{Sec_Conclusion}

We propose a novel QSViT methodology to systematically employ quantization for the Spiking Vision Transformer.
Our QSViT first investigates the impact of different quantization levels in different network layers, then identifies the base quantization settings for guiding the precision reduction. 
Afterward, QSViT performs guided quantization to select the appropriate quantization setting, which is then utilized for developing an efficient quantized network. 
The experimental results show that, our QSViT methodology saves memory footprint by 22.75\% and saves power consumption by 21.33\%, while preserving the accuracy within 2.1\% from the original non-quantized model on the ImageNet dataset, thereby demonstrating its potential to pave the way toward efficient SViT deployments on resource-constrained embedded AI systems. 

%%%%%%%%%%%%%%%%%%%%%%%%%%%%%%%%%%%%%%%%%%%%%%%%%%%%%%%%%%%%%%%%
%%%%%%%%%%%%%%%%%%%%%%%%%%%%%%%%%%%%%%%%%%%%%%%%%%%%%%%%%%%%%%%%
% \section*{Acknowledgment}
% ...

\bibliographystyle{IEEEtran}
\bibliography{bibliography.bib}
\end{document}